\documentclass[10pt,journal,compsoc]{IEEEtran}
\ifCLASSOPTIONcompsoc
  \usepackage[nocompress]{cite}
\else
  \usepackage{cite}
\fi
\ifCLASSINFOpdf
  \usepackage[pdftex]{graphicx}
  \graphicspath{{image/}{../jpeg/}}
  \DeclareGraphicsExtensions{.pdf,.jpeg,.png}
\else
\fi
\usepackage{amsmath}
\begin{document}
\title{A Taught-Obesrve-Ask (TOA) Method for Object Detection with 
Critical Supervision}
\author{Chi-Hao~Wu, Qin~Huang, Siyang~Li,
and~C.-C.~Jay~Kuo,~\IEEEmembership{Fellow,~IEEE}
}


\IEEEtitleabstractindextext{%

\begin{abstract}
Being inspired by child's learning experience - taught
first and followed by observation and questioning, we investigate a
critically supervised learning methodology for object detection in this
work. Specifically, we propose a taught-observe-ask (TOA) method that
consists of several novel components such as negative object proposal,
critical example mining, and machine-guided question-answer (QA)
labeling. To consider labeling time and performance jointly, new
evaluation methods are developed to compare the performance of the TOA
method, with the fully and weakly supervised learning methods.  Extensive
experiments are conducted on the PASCAL VOC and the Caltech benchmark
datasets.  The TOA method provides significantly improved performance of
weakly supervision yet demands only about 3-6\% of labeling time of full
supervision. The effectiveness of each novel component is also analyzed. 
\end{abstract}

}

\maketitle

\IEEEdisplaynontitleabstractindextext
%
\IEEEpeerreviewmaketitle

\IEEEraisesectionheading{\section{Introduction}\label{sec:introduction}}

\IEEEPARstart{T}{he} superior performance of convolutional neural
networks (CNNs) is attributed to the availability of large-scale
datasets with human labeled ground truth. While visual data are easy to
acquire, their labeling is time-consuming.  There exists a significant
gap between labeled and unlabeled data in real world applications.  To
address this gap, it is essential to develop weakly supervised solutions
that exploit a huge amount of unlabeled data and a small amount of
carefully selected labeled data to reduce the labeling effort. We
address this issue using the object detection task as an example, which
is one of the most fundamental problems in computer vision. 

Despite extensive studies on weakly supervised object detection in
recent years \cite{yan2017weakly,bilen2014weakly,gokberk2014multi,
bilen2015weakly, russakovsky2012object,song2014learning,song2014weakly,
siyang2017}, its performance is much lower than that of fully supervised
learning. Here, we examine this problem from a brand new angle - how to
achieve target performance while keeping the amount of labeled data to
the minimum. This is called the ``critically supervised" learning
methodology since only the most critical data are labeled for the
training purpose.  This idea is inspired by child's learning behavior.
Children are first taught to recognize objects by parents or teachers
given only a few examples.  Then, they keep observing the world and
start to ask questions. Sometimes, they ask YES/NO-oriented questions to
confirm their hypothesis.  These questions are from the most critical
examples that can enhance their understanding. 

It is worthwhile to take a further step to examine the gap between CNN
learning and human learning. For CNN-based object detection, large-scale
training datasets are needed to train a network with a tremendous number of parameters.
To get enough training samples, precise bounding boxes of objects are
provided. These fully labeled data are required to generate positive
(object) and negative (background) samples simultaneously in training.
In human learning, only little supervision is required for positive
samples. Besides, humans have innate ability to obtain background
samples either through tracking or 3D vision so that no additional
supervision is needed for negative samples.  

To reduce the labeling effort for CNNs, we explore the negative object
proposal (NOP) concept that collects a huge number of background samples
with little supervision. As to positive samples, we argue that it is
more natural and faster to apply the question-answer (QA) labeling than
drawing tight bounding boxes around objects.  According to
\cite{papadopoulos2016we}, it takes around 42 seconds to draw a
high-quality bounding box while it takes only 1.6 seconds for human to
do one QA verification.  To reduce the number of questions required to train
a high precision detector, a critical example mining (CEM) method is
proposed to select the most critical samples through QA.  These novel ideas are integrated to yield one
new solution called the taught-observe-ask (TOA) method since it mimics
how children learn to recognize objects. Being different from the weakly
supervised learning methods that get low precision from fewer labels,
the TOA method achieves better performance with less labeling time. 

\begin{figure}[h!]
\centering
\includegraphics[width=0.4\textwidth]{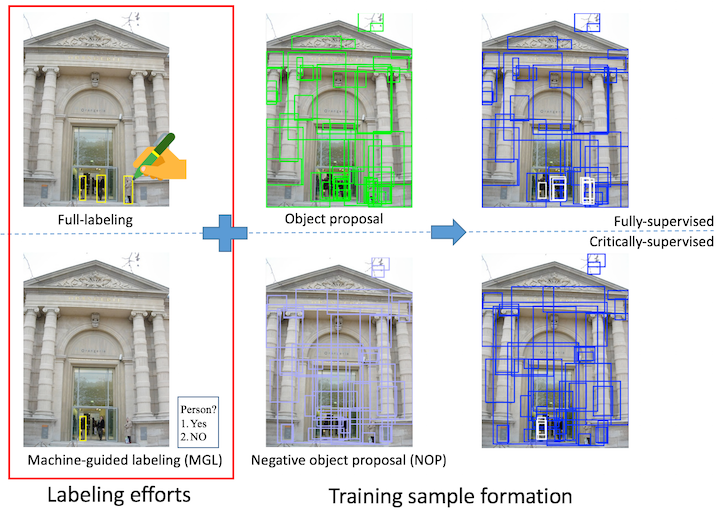}
\caption{Comparison of fully-supervised training (top) and
critically-supervised training (bottom).  All objects are labeled with
tight bouding boxes and training samples are obtained from calculating
their IoU with object proposals in full supervision.  In contrast,
critical examples and NOP samples are used in the training of a
critically supervised TOA method, where QA labeling is only needed for
critical examples selected by the CEM algorithm.}
\label{fig.TOA_illustrate}
\end{figure}

The fully supervised training and the critically supervised training
methodologies are compared in Fig. \ref{fig.TOA_illustrate}, where the
latter uses the TOA method as an example.  Both critical examples and
NOP samples are used in the training of a critically supervised TOA
method. To be more specific, the CNN model is first trained with a small
amount of fully labeled data. Then, extra examples are selected from the
set of unlabeled data by using the CEM algorithm. Finally, a series of
QAs are conducted for humans to verify labels. Afterwards, the CNN is
retrained using critical examples and NOP samples. 

There are several major contributions of this work. First, a critically
supervised learning methodology is explored to take data labeling and
learning into account jointly.  Second, the proposed TOA method contains
several new ideas such as the negative object proposal (NOP) and
critical example mining (CEM) that selects samples for labeling using a
dynamically varying rule.  Third, new evaluation methods are developed
to compare performance against labeling time.  Finally, extensive
experiments are provided to validate the training strategies and
effectiveness of each component in TOA.  The TOA method is evaluated on
the PASCAL VOC datasets \cite{Everingham15} for multi-class object
detection, and on the Caltech pedestrian dataset
\cite{dollar2012pedestrian} for single-class object detection as a
special case.  The TOA method provides an mAP (mean Average Precision)
improvement of 9.4\% over state-of-the-art weakly supervised detectors
under the same labeling time. As compared with the fully-supervised
learning method, the labeling time saving ranges from 66-95\% under the
same performance. 

\section{Related Previous Work} \label{sec.TOA_previousworks}

\subsection{Object Detection}

Obeject detection is one of the most intensively studied problems in
computer vision. It remains to be a challenging task. Deep learning has
brought great success to this area in recent years. Its solution
outperforms traditional methods such as the deformable part model (DPM)
\cite{felzenszwalb2010cascade,felzenszwalb2008discriminatively} by a
siginificant margin. The R-CNN \cite{girshick2014rich} accepts each
warped region as the CNN input for object classification. Region
proposals \cite{zitnick2014edge, uijlings2013selective} are needed in
the R-CNN. Other CNNs such as the Fast-RCNN \cite{girshick2015fast} and
SPP \cite{he2014spatial} accept the whole image as the input to the CONV
layers so as to reduce redundant convolutional computations. Their
region proposals are pooled in later layers for further classification
and bounding box regression. The Faster-RCNN \cite{ren2015faster}
further extends this idea by including a region proposal network to
avoid the need of object proposals outside the network. Recent
extensions \cite{kong2017ron,redmon2016you,dai2016r,kong2016hypernet,
liu2016ssd} either provide more efficient implementations with
competitive results or keep improving the mAP performance with more
advanced network design. 

\subsection{Weakly Supervised Learning}

Weakly supervised object detection is another active research topic
since it is not easy to collect a large amount of labeled data with
tight bounding boxes for the training purpose.  Given all object classes
in one image as the ground truth, weakly supervised object localization
(WSOL) techniques are designed to localize the object and enhance the
classification accuracy. There has been rapid progress along this line
using CNNs \cite{shi2017weakly,shi2016weakly,deselaers2012weakly,yan2017weakly,bilen2014weakly,gokberk2014multi,
bilen2015weakly,russakovsky2012object,song2014learning,song2014weakly,
siyang2017}. Most of them are based on multiple instance learning (MIL),
where images are treated as a bag of instances, and images that contain
no object instances of a certain class are labeled as negative samples
for this category, and vice versa. However, learning a detector without
the bounding box information is challenging. The performance of weakly
supervised detectors is significantly lower than that of fully
supervised detectors. 

\subsection{Active Learning and Others}\label{subsec:active}

Active learning is a technique that iteratively selects samples from a
large set of unlabeled data and asks humans to label so as to retrain
more powerful models. Previous active learning work mainly focuses on
tasks such as image classification \cite{joshi2009multi,
kapoor2007active, kovashka2011actively, qi2008two} and region labeling
\cite{siddiquie2010beyond, vijayanarasimhan2009multi,
vijayanarasimhan2009s}.  These methods usually start from a good
pretrained model and boost its performance with abundant extra unlabeled
data. Little work in active learning has been developed for window-based
object detection. Another related technique is called
``Human-in-the-loop labeling" that considers human-machine collaborative
annotation \cite{branson2010visual, vijayanarasimhan2014large,
siddiquie2010beyond}.  These methods are used when a pretrained model
does not provide satisfying performance on certain challenging tasks
and humans are asked to annotate more samples for performance
improvement.  Several other weakly supervised or unsupervised methods
\cite{liang2015towards,zhang2017mining,papadopoulos2017training} are also related to our work in
certain aspects. 

\begin{figure*}[h!]
\centering
\includegraphics[width=0.95\textwidth]{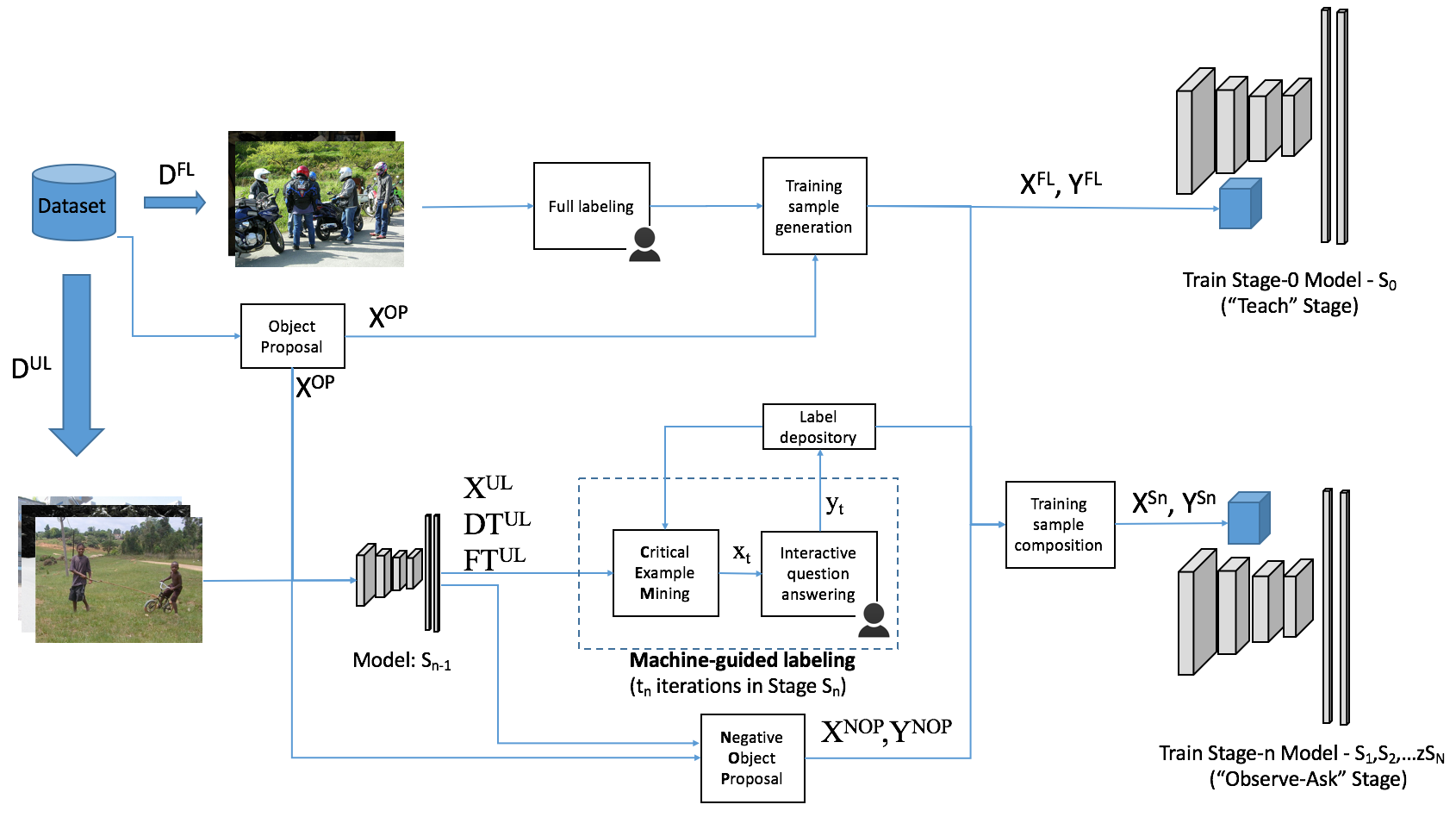}
\caption{The system diagram of the proposed TOA method.  The upper
branch represents the ``teach" stage where full labeling is conducted in
a small subset of images to train the initial model $S_0$. The lower
branch is the ``obeserve-ask" stage, where the CEM algorithm is used to
select critical examples from an unlabeled dataset using deep features
from model $S_{n-1}$ and the QA labeling is conducted on selected
examples.  The labeled examples are combined with NOP samples to form
the training set for stage-$n$ model $S_n$. Details are
given in Sec. \ref{sec.TOA_method}.}\label{fig.TOA_framework}. 
\end{figure*}

\section{Critically Supervised Learning}

People feed the CNN with a huge amount of labeled data in training, yet
some of them are not very critical. For example, fully supervised object
detection requires accurate bounding boxes of all target objects in one
image, which is time consuming and may not be optimal in striking a
balance between efforts and performance.  Being inspired by human
learning, CNN training can be achieved in a different way. That is, only
most relevant samples are labeled. By critically supervised learning, a
model is trained to reach a target performance with a minimum amount of
labeled samples. This is different from traditional weakly supervised
learning, which provides a limited number of labels (or label types) yet
does not pay attention to which parts of the dataset are critical to the
performance. 

Although critically supervised learning is related to active learning or
human-in-the-loop labeling as reviewed in Sec.  \ref{subsec:active}, it
has a clear objective; namely, minimizing the labeling time under a
target performance. To achieve the goal, we may adjust labeling
strategies dynamically in different labeling stages to ensure that any
new labeling effort helps boost the performance.  To take object
detection as an example, a CNN needs a large number of training samples
to gain superior performance while most of them are background (or
negative) samples. The aquisition of background samples relies on
accurate labeling of all bounding boxes to avoid false negatives.  We
will present a new way to provide negative samples more effectively.
Furthermore, not all object samples share the same importance. We will
show that only a small number of them are representative and critical
for a CNN to gain discriminative power in separating different classes.
The adoption of various labeling strategies and mining criteria is the
key to the success of critically supervised learning. 

\section{Taught-Observe-Ask (TOA) Method} \label{sec.TOA_method}

\subsection{System Overview}

Being inspired by children learning, we propose the TOA method to
achieve critically supervised learning.  The overall system diagram of
the TOA method is shown in Fig. \ref{fig.TOA_framework}. We split a
large amount of unlabeled images ($D$) into two subsets.  They are sets
of fully labeled and unlabeled images, denoted by $D^{FL}$ and $D^{UL}$,
respectively. Subset $D^{FL}$ consists of only a small number of fully
labeled images.  Training samples $(X^{FL},Y^{FL})$ are generated from
them and used to train a stage-$0$ detector, $S_0$, as shown in the
upper branch of Fig.  \ref{fig.TOA_framework}.  This initial stage,
called the ``teach" stage, is needed for the machine to understand the
problem definition, gain basic recognition capability and know what
questions to ask in later stages. 

\begin{figure}[h!]
\centering
\includegraphics[width=0.5\textwidth]{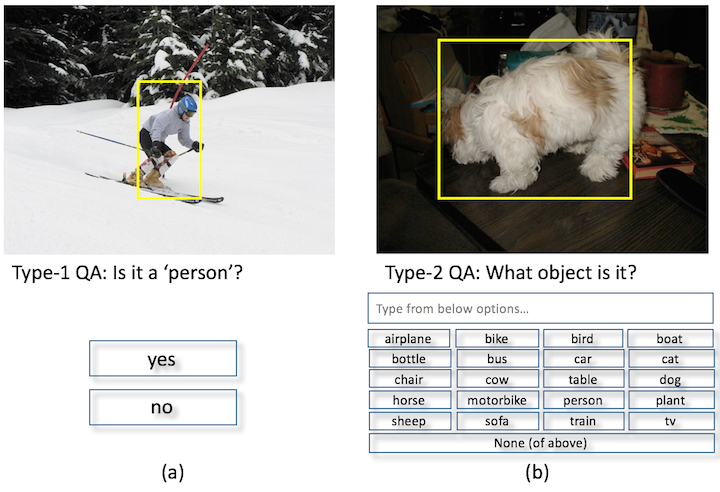}
\caption{Illustration of two QA types: (a) type-1 QA and (b) type-2
QA.}\label{fig.TOA_qa_illustrate}
\end{figure}

All remaining stages are the ``observe-ask" stage.  In stage $n$, $n=1,
2, \cdots$, we collect necessary features $FT^{UL}$ and detection scores
$DT^{UL}$ for each region proposal using the CNN model trained in the
previous stage. The CEM module uses features to mine the most critical
example, $x_t$, from all unlabeled samples and asks humans to verify (or
determine) its label $y_t$ (see Fig. \ref{fig.TOA_qa_illustrate}).
Then, we gather $t_n$ labeled samples through $t_n$ iterations, and
combine them with negative object proposals, denoted by $X^{NOP}$, to
form a training set of sample/label pairs $(X^{S_n},Y^{S_n})$.  A new CNN
model, $S_n$, is then trained using all labeled data up to stage $n$. We
summarized used notations in Table \ref{tab.TOA_notation}.  In Fig.
\ref{fig.TOA_framework}, the machine-guided labeling (MGL) is used to
find the most critical example and conduct question-answer (QA) labeling to
obtain labels from humans iteratively.  The NOP is used to collect
background samples without extra supervision. 

\begin{table}[]\centering
\caption{Summary of notations used in this work. Note that $N^{OP}_i$
denotes the number of object proposals in image $i$ and $N^{OP'}$ is
different from $N^{OP}$ when only a subeset of object proposals are used
in the training. $L = \{1...(N+1)\}$, where $N+1$ is the total number of
object classes and the background is included as an extra class.}
\label{tab.TOA_notation}
\resizebox{0.5\textwidth}{!}{
\begin{tabular}{|l|l|}
\hline
Number of images  &   $N^D$ \\ \hline
Dataset &   $D = \{d_i \in Z^{w_i \times y_i} | i =1...N^{D}\}$ \\ \hline 
Object proposal &  $X^{OP} = \{x_{i,j} \in Z^4 | i=1...N^D, j=1...N^{OP}_i\}$ \\ \hline
        & Fully labeled subset \\ \hline
Number of images  &   $N^{FL}$   \\ \hline
Image index   &   $\Phi^{FL} = \{\phi_i \in Z | i=1...N^{FL}\}$   \\ \hline
Image subset  &   $D^{FL} = \{d_i \in Z^{w_i \times y_i} | i \in \Phi^{FL}\}  $ \\ \hline
Samples   &   $X^{FL} = \{x_i \in Z^4 | i \in \Phi^{FL}, j=1...N^{OP'}_i\}$ \\ \hline
Labels    &   $Y^{FL} = \{y_i \in L | i \in \Phi^{FL}, j=1...N^{OP'}_i\}$  \\ \hline
        & Unlabeled subset \\ \hline
Number of images   & $N^{UL}$ \\ \hline
Image index  &  $\Phi^{UL} = \{\phi_i \in Z | i=1...N^{UL}\}$ \\ \hline
Image subset & $D^{UL} = \{d_i \in Z^{w_i \times y_i} | i \in \Phi^{UL}\}  $ \\ \hline
Samples    & $X^{UL} = \{x_i \in Z^4 | i \in \Phi^{UL}, j=1...N^{OP'}_i\}$\\ \hline
Labels    &  $Y^{UL} = \{y_i \in L | i \in \Phi^{UL}, j=1...N^{OP'}_i\}$ \\ \hline
Deep Feature     & $FT^{UL} = \{ft_i \in R^{dim(ft)} | i \in \Phi^{UL}, j=1...N^{OP'}_i\}$ \\ \hline
Detection score   & $DT^{UL} = \{dt_i \in R^{N+1} | i \in \Phi^{UL}, j=1...N^{OP'}_i\}$  \\ \hline
\end{tabular}}
\end{table}

\subsection{Machine-Guided Question-Answer Labeling}

Annotating tight bounding boxes (i.e. full labeling) is a time-consuming
and demanding job.  It is easier to ask humans to answer whether an
object class exists in an image yet without the bounding box information
(i.e. weak labeling).  The drawback of the latter is that its detection
performance is lower. In this work, we adopt another labeling scheme,
called question-answer (QA) labeling, which can be easily done yet with
the bounding box information. We implement two QA labeling types as
shown in Fig.  \ref{fig.TOA_qa_illustrate}. For the first one, humans
answer the yes-or-no question to verify whether a sample belongs to a
certain object class. For the second one, humans choose an object class
from a list. They are refered to as type-1 and type-2 QA, respectively.
Consecutive type-1 QAs are as effective as a type-2 QA, but the labeling
time can be reduced with type-1 QA when we are confident about the
object class of the sample. 

The TOA method has one more advantage.  To make training effective,
humans need to label all objects in one image without missing any
objects in fully or weakly supervised schemes. Failing to do so will
result in performance drop. Thus, extra efforts are required for
verification. The need to label all objects is eased by the adoption of
NOP in the TOA method. In other words, annotators only need to focus on
one specific sample at a time. 

It is, however, not always possible for the machine to select samples
with perfectly tight bounding boxes for QA. In the QA-labeling process, we require
humans to give answers with tolerance to noisy samples (i.e. samples
without perfect bounding boxes). A guideline is provided to annotators.
For example, they should accept objects when the sample overlaps with
the drawn bounding box with the intersection-over-union (IOU) over a
certain threshold value (e.g., 0.6).  A few examples that meet the
threshold values are provided to annotators in the beginning of the
annotation process. Humans can learn such a rule quickly.  Although
small variations may still occur when the overlapping degree is close to
0.6, the final performance is not much affected as shown in Sec.
\ref{sec.TOA_results}. 

\subsection{Negative Object Proposal (NOP)}

The object proposal \cite{zitnick2014edge,uijlings2013selective} is a
technique to extract a set of bounding boxes that ideally include all
objects in one image. It reduces the computational complexity of
exhaustive search. Here, we present a scheme, called the negative object
proposal (NOP), that collects a set of boxes that contain no objects.
With the NOP, the need of labeling all objects in one image to collect
negative samples is eliminated.

\begin{figure}[h!]
\centering
\includegraphics[width=0.5\textwidth]{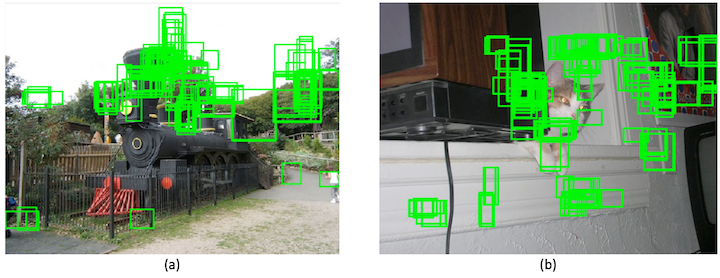}
\caption{Examples to illustrate the challenges in extracting good NOP
samples, where only 200 out of all boxes are plotted for ease of
visualization: (a) edge-box results with extremely low scores which rank
over 2000 in one image and (b) the edge-box results with the modified
formula.}\label{fig.TOA_nop_failure}
\end{figure}

Distinguishing objects from non-objects appears to be effortless for
humans, yet unsupervised NOP extraction is not a trivial job.  To show
the challenge, we examine two ways to extract the NOP.  First, we
collect boxes with extremely low scores from an object proposal
algorithm called the EdgeBoxes \cite{zitnick2014edge}. Second, we modify
the formula of EdgeBoxes to weigh more on boxes that have more edges
going across box boundaries. The results are shown in Fig.
\ref{fig.TOA_nop_failure}.  We see two major problems in these results.
First, the algorithms tend to contain many small bounding boxes. To
train a CNN, the NOP should contain some informative samples of
reasonable sizes and overlap with objects to a certain degree.  Second,
reasonable training performance could only be achieved with high NOP accuracy 
since negative samples are the majority in the training
samples. However, the boxes generated by these algorithms occasionally
contain positive samples and fail to meet the required precision. 

To address these challenges, we use the detection score information
$DT^{UL}$ from the stage-0 detector only. In other words, we do not need
extra labeling efforts for the NOP. Although the stage-0 model is
trained by a small percentage of data and its detection scores are not
robust, one can still obtain a candidate set of samples that contain
almost all objects with a sufficiently low threshold. Then, NOP samples
can be extracted by avoiding samples that are spatially close to the
candidate set. This idea can be formalized mathematically below.

We first define an inverse set concept based on the IoU, where the
function returns samples in set A that do not have a large overlap
degree with any samples in set B:
\begin{equation}
C = \mbox{InvSet} (A,B) = \{a_i \in A | IoU(a_i, b_j) < \tau, \forall b_j \in B \}.
\end{equation}
With Eq. (\ref{eq.ch5_inv_iou}), the NOP set can be calculated by finding
the inverse set from a candidate object set that is extracted using
detection scores. It can be written as
\begin{equation}\label{eq.ch5_inv_iou}
X_i^{NOP} = \mbox{InvSet} (X_i^{OP}, X_i^{DT_\epsilon}), \quad \forall i \in \Phi^{UL},
\end{equation}
where $X^{DT_\epsilon}$ is the candidate set fetched by applying
threshold $\epsilon$ to detection scores and $X^{OP}$ contains samples
generated by an object proposal algorithm. The resulting NOP set,
$X^{NOP}$, not only provides high quality negative samples of reasonable sizes but also has a very low probability to include object
samples. Experiments will be conducted in Sec.  \ref{sec.TOA_analysis}
to support such a claim. 

It is sometimes possible to create the NOP samples without any supervision (e.g. without the stage-0 detector) if we
have prior knowledge about target objects. For example, for the
pedestrian detection problem, almost all pedestrians are in the upright
direction with a certain range of aspect ratios. We can simply make the
candidate object set contain boxes whose shapes are in the range of all
possible pedestrian aspect ratios. The NOP calculated from Eq.
(\ref{eq.ch5_inv_iou}) will generate boxes of wrong pedestrian aspect
ratios. This is used in experiments in Sec. \ref{sec.TOA_results}, and
we obtain good performance even if it is totally unsupervised. 

\subsection{Critical Example Mining (CEM)}

\begin{figure}[h!]
\centering
\includegraphics[width=0.5\textwidth]{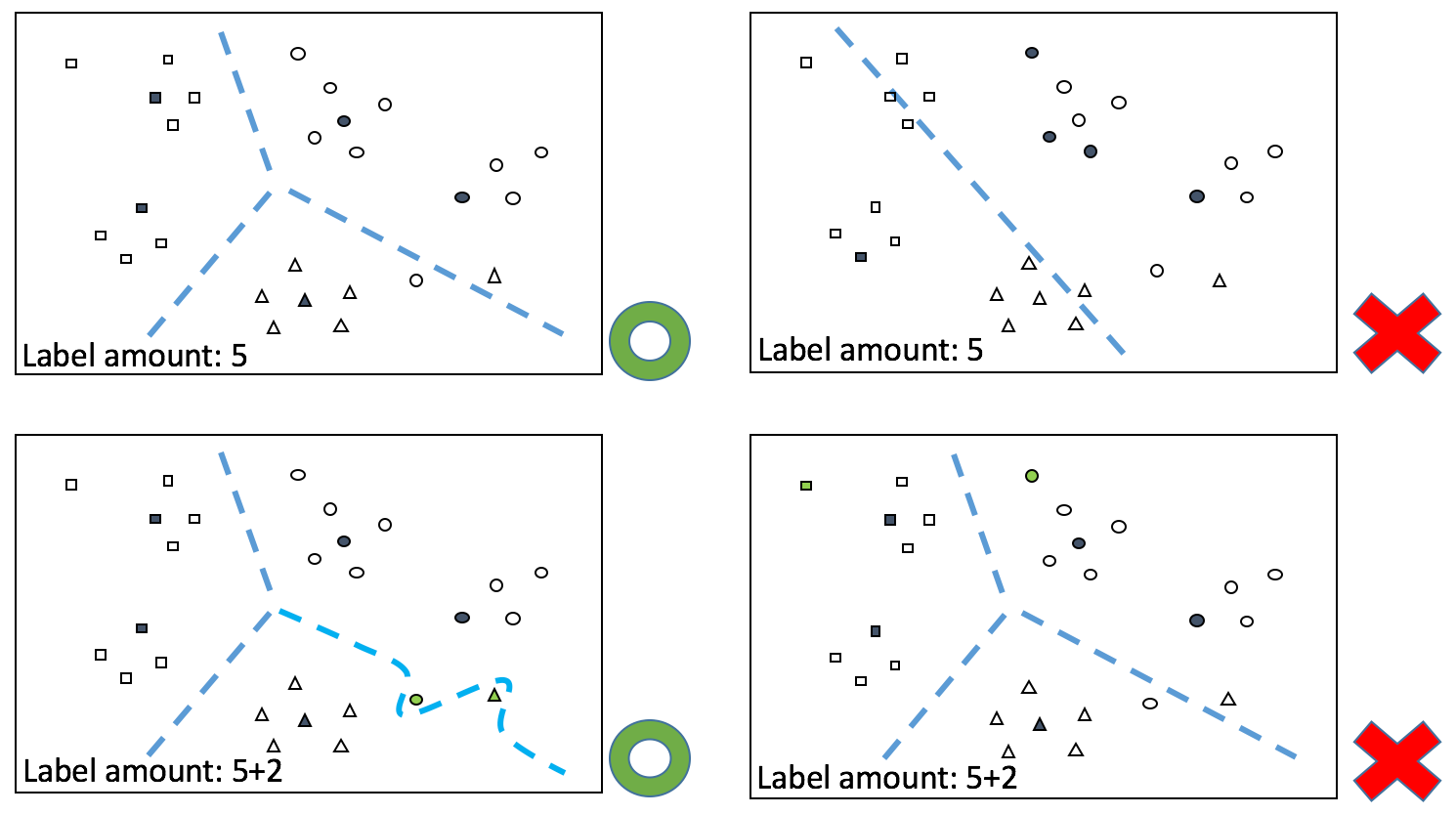}
\caption{Illustration of the critically-supervised learning methodology,
where solid samples are labeled samples while other are unlabeled ones.
Top) Under the constraint of using only 5 QAs, selection of samples for
labeling has major impact on the trained detector (classifier). The left
choice is good while the right choice is poor.  Bottom) When additional
2 samples (in green) can be selected to achieve better performance, the
selection rule is changed.  Again, the left choice is good while the
right choice is poor.} \label{fig.TOA_cem_concept}
\end{figure}

It is well known that some samples play more important roles than others
in the CNN training as elaborated in \cite{shrivastava2016training,
chihao2017}. Typically, hard samples are more critical to performance
improvement. In critically supervised learning, we claim that the
criteria of choosing important samples should change dynamically along
the labeling process, and use Fig. \ref{fig.TOA_cem_concept} to support
this claim.  For a limited labeling budget, the selection of examples
should be cautious so as not to waste any labeling effort. Initially,
two criteria are more important: 1) the selection should be balanced
across different classes; and 2) the sample should be representative, and
redundant ones should be avoided. As shown in the top row of Fig.
\ref{fig.TOA_cem_concept}, cautious selection of labeled samples using
these two criteria is adequate.  As more samples are labeled and the
model starts to gain good performance on easy samples, it becomes
important to find harder examples to achieve better separation in
decision boundaries as shown in the bottom row of Fig.
\ref{fig.TOA_cem_concept}. This type of CEM is a new problem that has never been
addressed before, to the best of our knowledge.

We define a function, $c_{i,j}$, to capture the criticalness of each
sample by taking three components into consideration. They are class
balancing (BAL), sample representativeness (REP), and hardness (HARD). 
The criticalness function can be written as the sum of these three
components:
\begin{equation}
c_{i,j} = c_{i,j}^{BAL} + c_{i,j}^{REP} + c_{i,j}^{HARD}, 
\end{equation}
where
\begin{eqnarray}
c_{i,j}^{BAL} & = & cs_{i,j}^{BAL} \cdot cp_{i,j}^{BAL}(t), \\
c_{i,j}^{REP} & = & cs_{i,j}^{REP} \cdot cp_{i,j}^{REP}(t), \\
c_{i,j}^{HARD}& = & cs_{i,j}^{HARD} \cdot cp_{i,j}^{HARD}(t).
\end{eqnarray}
Since the importance of each component varies along the labeling
process, we express it as the product of two terms. The first term,
$cs_{i,j}$, represents its score function while the second term, $cp_{i,j}$,
called the progress function, controls the varying contribution of each
component in the $t$th iteration. There is no ground truth about
importance of each sample along the labeling process, and this function
is chosen heuristicly. Before explaining each component, we examine the
characteristics of the criticalness function. 

\noindent{\bf Detection Score and Deep Features.} The detection score,
$dt_{i,j}$, is the output of the CNN that represents the probability of
a sample belonging to a specific class. They are obtained in each
stage-$n$ by feeding all training samples into the $S_{n-1}$ model for
prediction.  Although the precision of $S_{n-1}$ may not be high enough,
its detection score of a sample still serves as an indicator of how
likely the sample could be an object. Each score $dt_{i,j}$ is a
vector with a dimension of $(N+1)$ (i.e., $N$ object classes plus one
background class).  Besides detection scores, deep features can be
extracted from CNN prediction and they provide useful information in
mining critical examples. Deap features, $FT$, are the output from one
of the intermediate CNN layers and the feature dimension depends on
which layer in use.  Detailed experimental set-ups will be provided in
Sec.  \ref{sec.TOA_results}. 

\begin{figure}[h!]
\centering
\includegraphics[width=0.5\textwidth]{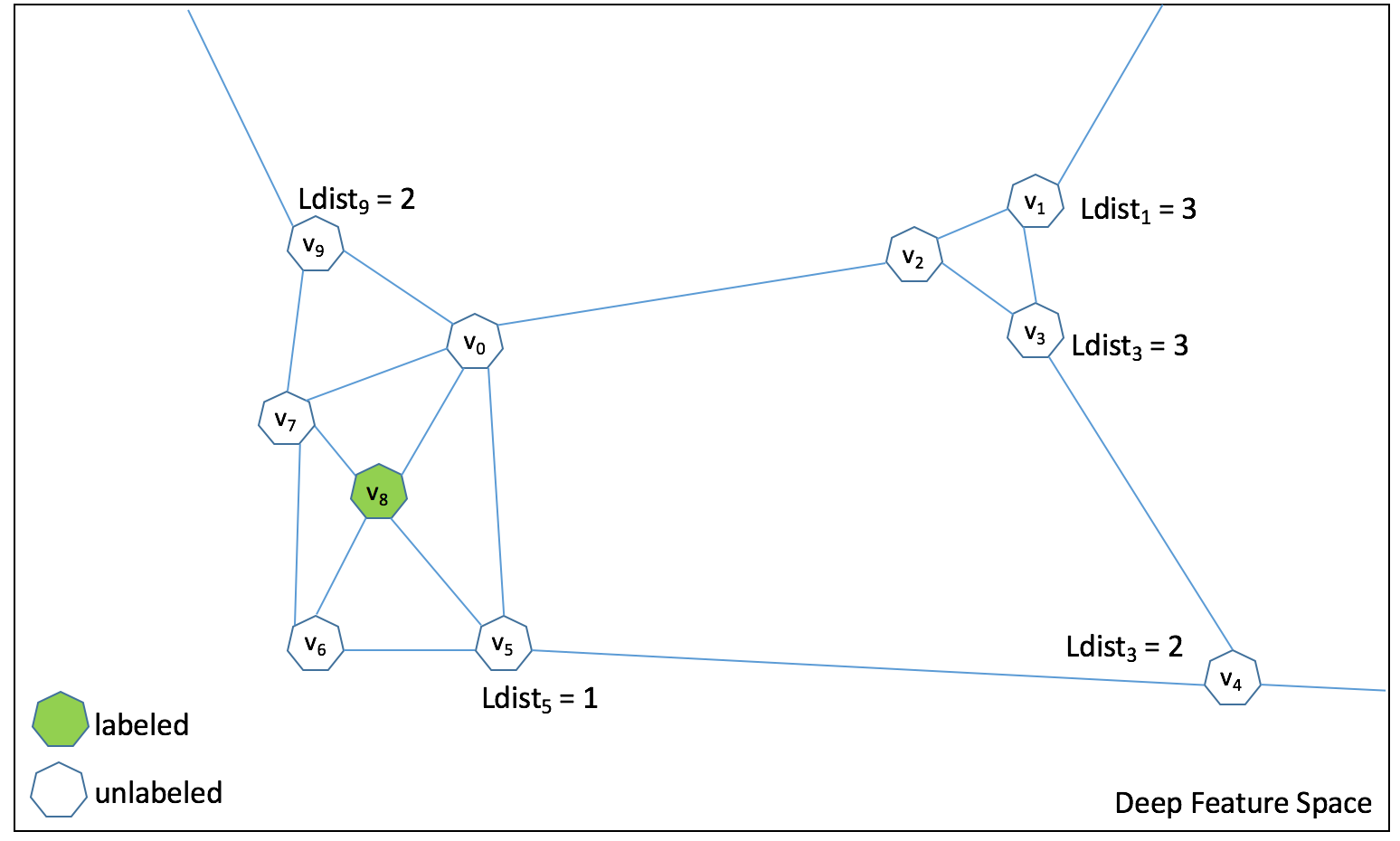}
\caption{Illustration of the kNN graph where each node is connected to its 
$K$ nearest neighbors (K=3 in this example). The Ldist of a node is 
the number of edges along the shortest path to its nearest labeled
node.}\label{fig.TOA_knngraph}
\end{figure}

\noindent 
{\bf KNN Graph and Geodesic Distance.} To measure the representativeness
of a sample, we design our mining algorithm based on the
k-nearest-neighbor (kNN) graph. To build the undirected kNN graph
$G=(V,E)$, each node is a sample in either $X^{UL}$ or $X^{FL}$. We
calculate the Euclidean distances of each node $v_i$ with other samples
in the deep feature space, which is denoted by $d(v_i,V)$. Sample $v_2$
is set to connect with $v_1$ in the graph if it is among the $k$
nearest neighbors of $v_1$. One can draw the kNN graph based on the
following formula:
\begin{equation}\label{eq.ch5_graph}
e_{v_1,v_2} = \left\{\begin{matrix}
1, & d(v_1,v_2) \le \mbox{ascend-sort}(d(v_1,V))[K], \\ 
1, & d(v_1,v_2) \le \mbox{ascend-sort}(d(v_2,V))[K], \\
0, & \mbox{otherwise}.
\end{matrix}\right.
\end{equation}
With the kNN graph, the geodesic distance between two nodes is the
shortest path between two nodes in the kNN graph \cite{zhang2017mining}.
As compared with the Euclidean distance, the geodesic distance better
represents the sample distribution over the deep feature space.  For
each unlabeled sample, we define a metric, $ldist_{i,j}$, to represent
the nearest labeled sample measured by the geodesic distance.  When
$ldist_{i,j}$ is smaller, sample $x_{i,j}$ is less critical since there
is a labeled sample nearby.  An exemplary kNN graph and its associated
geodesic distance and DLIST metric are shown in Fig. \ref{fig.TOA_knngraph}. 

\noindent
{\bf Progress Function.} To control the contribution of each component in
the criticalness function dynamically, we define a simple progress
function below:
\begin{equation}\label{eq.ch5_progress}
P(x,\mu,\sigma^2) = \left\{\begin{matrix}
1, & x < \mu, \\ 
N(x | \mu, \sigma^2) / N(\mu | \mu, \sigma^2), \quad & x \ge \mu,
\end{matrix}\right.
\end{equation}
where $N(x|\mu,\sigma^2)$ is the Gaussian with mean $\mu$ and variance
$\sigma^2$.  A component is set to be important in the labeling process
until a certain condition is met at threshold $\mu$. After that, we
reduce its weight gradually in Gaussian decay form. Then, we can move
the focus to other criteria in critical examples selection.

\noindent

{\bf Criticalness Function.} We elaborate on the three components of the
criticalness function. The sample balancing (BAL) component considers
the balance between different object classes and the balance between
positive and negative samples simultaneously. In earlier labeling
stages, there are much more negative samples obtained by the NOP
algorithm.  Selecting a sample with a higher probability of positive
samples is important so that the QA effort is not wasted.  Hence, the
maximum detection score, $\max(dt_{i,j})$, of a sample serves as a
good index in sample selection; namely,
{\small
\begin{eqnarray}\label{eq.ch5_balance}
c_{i,j}^{BAL} = \max(dt_{i,j})\cdot P(\mbox{count}(y_{i,j} > 0, \forall y_{i,j} 
\in Y^{UL}_{i}), (\mu,\sigma^2)^{BAL1}) \nonumber \\
\cdot P(\mbox{count}(y_{i,j} = \mbox{argmax}(dt_{i,j}), \forall y_{i,j} 
\in Y^{UL}), (\mu,\sigma^2)^{BAL2}).  \nonumber
\end{eqnarray}}
This contains two progress functions. The first one is to lower BAL's
importance when more object samples are labeled. The second one is to
ensure the balance between different classes. This is achieved by
setting a threshold at which a specific class already has enough labeled
samples in the current stage. For instance, we can avoid selecting too
many examples from major classes such as ``people" or ``car" in the VOC
dataset and give the QA quota to other rare classes to boost the overall
performance in earlier stages. 

The sample representativeness (REP) component considers the
representativeness of a sample.  As discussed earlier, when
$dlist_{i,j}$ is smaller, the sample is relatively redundant in
training, and vice versa. Thus, we have
\begin{equation}\label{eq.ch5_representative}
c_{i,j}^{REP} = \delta \cdot -P(ldist_{i,j}, (\mu,\sigma^2)^{REP}).
\end{equation}
where $DLIST$ is in the progress function since its values may keep
changing in the labeling process. Besides the progress function, we use
variable $\delta$ to adjust the importance of this component across
different datasets. Some datasets contain much more redundant samples
while others not.  The choice of $\delta$ is determined by the sample
distribution in the kNN graph and will be further discussed in Sec.
\ref{sec.TOA_analysis}. 

The hardness (HARD) component gives a higher priority to hard samples
which appear in decision boundaries. This component is not critical
until many representative samples are labeled. We define
\begin{eqnarray}\label{eq.ch5_hardness}
c_{i,j}^{HARD} & = & \left( \sum(dt_{i,j}) - \max(dt_{i,j}) \right) 
\cdot \nonumber \\ && -P(\mbox{mean}(LDIST), (\mu,\sigma^2)^{HARD}),
\end{eqnarray}
where the progress function is determined by the average $DLIST$ value
since it is an indicator of how densely samples are labeled in the deep
feature space. This component is important only when labeled
samples are dense enough in the feature space.  To determine the hardness of a sample, we 
examine the detection scores and check if they are high not only in one but in several
object classes. 

The criticalness score is used to determine the importance of a sample.
The CEM procedure computes the criticalness scores of all unlabeled
samples, and the sample with the highest score is selected for the QA
labeling. In general, we apply the type-1 QA labeling by asking humans
whether the sample belongs to the class that has the highest score. For
hard samples which satisfies $cp_{i,j}^{HARD} > cp_{i,j}^{BAL}$ and
$cp_{i,j}^{HARD} > cp_{i,j}^{REP}$, the type-2 QA is used since we are
less sure about its class. The newly labeled sample is used to update
$dlist_{i,j}$ and $c_{i,j}$ values of affected unlabeled samples.  The
iteration continues until the end of the stage. 

\subsection{Training Sample Composition}

In TOA, we obtain some QA-labeled samples and NOP samples from
unlabeled images in each stage. To train the model, positive samples are chosen among
object proposals that have IoU greater than a threshold $TH_{FG}$ with QA-labeled positive samples.
Negative samples are selected from the NOP set with its IoU in the range
of $[TH_{HI}, TH_{LO}]$ with QA-labeled positive samples. The three thresholds
are similar to that used in fully supervised learning.  The differences
lie in that we use QA-labeled samples to calculate the IoU value and
determine positive/negative samples rather than the ground truth and
that we use the NOP instead of object proposals for negative samples.
Examples will be shown in Sec. \ref{sec.TOA_analysis}.  After the QA
number reaches the target amount in stage $n$, the newly labeled images
associated with the QA-labeled samples as well as all previous labeled
data are used to train CNN model $S_n$. 

It is worthwhile to point out that QA-labeled samples can be noisy (i.e.
the bounding boxes may not be tight). The composition of samples in the
TOA method with three thresholds may cause positive samples to have less
overlapping with the real ground truth and/or negative samples to have
more overlapping with the ground truth than the fully supervised case.
We will elaborate on the corresponding IoU distributions in Sec.
\ref{sec.TOA_analysis}. 

\section{Evaluation Method}

\subsection{Labeling Time Model} \label{sec.ch5_labeltime}

Human labeling time has been discussed in prior literature. For example,
it was mentioned in \cite{papadopoulos2016we} that drawing a tight
bounding box with high quality takes takes around 42 seconds in crowd
sourcing, and 26 seconds for faster annotation.  In contrast, the
labeling time for a simple yes-or-no question only requires 1.6 seconds.
Similar numbers were reported in \cite{su2012crowdsourcing,
papadopoulos2017extreme}. 

In order to evaluate the performance, we need to build a labeling time
model. Here, we use the numbers mentioned in these papers for bounding
box drawing and yes-or-no questions. In addition, we need the labeling
time for answering a type-2 question ({\em i.e.}, what class an object
belongs to). We conduct experiments on our own, and calculate the
average time needed to type a specific object class in multiple trials.
It is 2.4 seconds. Similarly, we evaluate the time for humans to scan
the whole image to ensure no missing objects, which is 2.6 seconds.
These four numbers are summarized in Table \ref{tab.TOA_time_model}
under the column of the high-quality (HQ) profile.  We also provide the
moderate-quality (MQ) profile in Table \ref{tab.TOA_time_model}.  The MQ
profile allows quicker drawing of bounding boxes and assumes that image
verification is done in parallel while drawing boxes. Experimental
results in Sec.  \ref{sec.TOA_analysis} demonstrate that the proposed
TOA method is competitive in both profiles. 

In this table, we also use a variable to denote a specific labeling time
to accommodate different labeling models since powerful tools can be
designed for more efficient annotation. For example, we can show a bunch
of samples together and make humans click on those that do not belong to
a certain class. This batch process will make the yes-or-no question
faster than 1.6 seconds per image in average. 

\begin{table}[h!]\centering
\caption{The labeling time model with different labeling types, where
the HQ profile is used for high-quality labeling and the MQ profile is
for moderate quality labeling.} \label{tab.TOA_time_model}
\begin{tabular}{|l|l|l|l|} \hline
Labeling type   & variable & HQ profile & MQ profile \\ \hline
Draw a tight bounding box &  $t^{FL}$  & 42.0 sec & 26.0 sec     \\ \hline
Answer type-1 question &  $t^{QA1}$ &  1.6 sec  & 1.6 sec     \\ \hline
Answer type-2 question &  $t^{QA2}$ &  2.4 sec  & 2.4 sec    \\ \hline
Verify missing objects &  $t^{VER}$ &  2.6 sec  & 0.0 sec  \\ \hline
\end{tabular}
\end{table}

\subsection{Estimated Labeling Time Ratio (ELTR)}

Given the labeling time model, we define a quantity, called the
estimated labeling time ratio (ELTR), to measure the overall performance
as compared to the fully or weakly supervised schemes. The ELTR
calculates the labeling time needed for a specific method normalized by
the time required for fully labeling for the whole image set.  We
provide detailed calculation of ELTR for different training scenarios
below. 

If we use a subset of images, $D^{FL}$, in fully supervised training,
its ELTR is the ratio between the number of selected image and the whole
dataset:
\begin{equation}\label{eq.ch5_full}
ELTR^{FS} = N^{FL} / N^{D}.
\end{equation}
For weak supervision, humans need to answer object classes in one image.
This is the same as the type-2 question answer. The difference is that
humans have to answer all object classes in one image in weak
supervision. Suppose that there are $N_i^{CLS}$ classes in image $i$, it
takes $N_i^{CLS}$ times $t^{QA2}$ seconds for weak supervision for this
image. Moreover, verification on the whole image has to be done in
weakly supervised learning. Thus, we have 
\begin{equation}\label{eq.ch5_weak}
ELTR^{WS} = \frac{\sum_{i=1}^{N^D}(t^{VER}+t^{QA2} N_i^{CLS})}{\sum_{i=1}^{N^D} 
t^{FL} \cdot N_i^{OBJ}},
\end{equation}
where $N_i^{OBJ}$ denotes the number of object presented in image $i$.
The labeling time is normalized by the time needed for full labeling of
the whole dataset. For critical supervision, our TOA method includes
full-labeling, and type-1 and type-2 QA labeling. All of them are added
up to yield
\begin{equation}\label{eq.ch5_critical}
{\small
ELTR^{CS} = \frac{\sum_{i \in \Phi^{FL}}t^{FL} N_i^{OBJ} + t^{QA1} N^{QA1} 
+ t^{QA2} N^{QA2}}{\sum_{i=1}^{N^D}t^{FL} N_i^{OBJ}}
.}
\end{equation}
where $N^{QA}$ is the total QA number and $N^{QA1}$ and $N^{QA2}$ are
the total QA numbers for type-1 and type-2 questions, respectively. 

\section{Experimental Results} \label{sec.TOA_results}

\subsection{Datasets}

To validate the proposed TOA method, we first test our algorithm on the
general multi-class object detection problem using the VOC 2007 and the
VOC 2012 dataset (abbreviated as VOC07 and VOC12, respectively). To
further demonstrate the generalization capability of the TOA method, we
conduct experiments on the Caltech pedestrian dataset (abbreviated as
the Caltech dataset), which is a special case of object detection with
only one object class. 

\noindent
{\bf VOC Dataset.} The VOC07 and the VOC12 datasets are two most widely
used object detection datasets. They provide 20 commonly seen object
classes. The VOC datasets are split into three subsets - train, val, and
test sets. In our work, we follow the standard procedure to train our
model on the trainval set and calculate the mean average precision (mAP)
on the test set. The VOC07 dataset has a total of 5011 trainval images
while the VOC07+12 dataset has a total of 16553 trainval images. 

\noindent
{\bf Caltech Dataset.} The Caltech dataset is the mostly used pedestrian
detection dataset. It provides a large number of training images
captured from the street with wide diversity. Pedestrian detection is
essential to various applications such as autonomous vehicles, the
advanced driving assistant system (ADAS), security, robotics, and
others. It is a special case of object detection which focuses on one
object class only.  However, it has several challenges such as cluttered
background, occlusion, extremely small objects, etc.  The Caltech
dataset provides a total of 128419 training images with their bounding
box ground truth. 

\subsection{Experimental Setup}

\noindent 
{\bf Simulation.} The TOA method involves iterative human-machine
collaborative labeling. To provide fair comparison with fully supervised
and weakly supervised learning methods, we choose to use existing
datasets that are already annotated. However, we do not use their labels
directly. Instead, we simulate our results by applying the QA-labeling
and let the ground truth in the dataset to answer our questions. For
example, when we choose a sample and ask a type-1 question, we assume
that humans will give the answer ``yes" when the sample has high enough
IoU with the ground truth, and the class that we query fits what is
annotated in the ground truth. For type-2 questions, we are expected to
get the object classes when IoU with the ground truth is greater than a
threshold. By doing the simulation, no human labeling is really needed.
In our work, the IoU threshold is set to 0.6.  We will provide more
analysis on the variation of the IoU threshold in Sec.
\ref{sec.TOA_analysis}. 

\noindent 
{\bf Network Architecture.} The evolution of CNN architecture is very
fast nowadays, and so is the CNN performance improvement on different
problems. In this work, we adopt the Fast-RCNN \cite{girshick2015fast}
(abbreviated as the FRCNN) as the baseline method for the proof of
concept.  The FRCNN does not provide state-of-the-art object detection
performance. We adopt it for two reasons. First, weakly supervised
object detection methods are still mostly based on FRCNN so that we use
it for fair comparison. Second, the FRCNN is a well-known method in
computer vision.  Considering that few CNNs have been tested on both the
Caltech and the VOC datasets, the FRCNN is a good choice
\cite{li2015scale, girshick2015fast}. Here, we use the VGG-16
\cite{simonyan2014very} as the back-bone network pretrained by the
ImageNet dataset. 

\noindent
{\bf Implementation Details.} In stage-$0$, a subset of images is
selected for full supervision. This subset is randomly selected from the
whole training set in the setup. Note that we tried different random
sets and the results do not change much. For object proposals, we adopt
the Edgeboxes \cite{zitnick2014edge} algorithm with its default
settings.  In the OA-stage, we select the responses at the FC-7 layer of
the CNN as the deep features denoted by $FT$. The FC-7 is the last layer
of the VGG-16 network. Its feature space contains multiple clusters 
that fit the TOA method well.

To determine detection scores and deep features for unlabeled training
images, the non-maxima-suppression (NMS) technique is adopted to reduce
redundant samples. In generating the kNN graph, each sample is computed
its Euclidean distance with all other samples, and the computational
overhead is high. To accelerate the process, we exclude samples with
detections scores lower than 0.01.  According to the designed CEM, these
samples are almost impossible to be selected. Furthermore, we generate
the kNN graph using the CUDA to exploit the GPU parallelism, which is
about 200 times faster than the CPU implementation. We set $K=4$ for the
kNN graph. 

Other parameters are empirically set as $\tau=0.3$, $\epsilon=0.01$,
$(\mu,\sigma^2)^{BAL1} = (1,0.2)$, $(\mu,\sigma^2)^{BAL2} =
(\#QA/20,1)$, $(\mu,\sigma^2)^{REP} = (1.5,0.5)$, and
$(\mu,\sigma^2)^{HARD} = (2,0.2)$, $(TH_{FG},TH_{HI},TH_{LO}) = (0.6,0.4,0.1)$. Other training options (e.g. REG, PRETR, WTR)
will be discussed later in Sec. \ref{sec.TOA_details}. 
 
\subsection{Multi-Class Object Detection}\label{sec.TOA_VOC}

\begin{figure}[h!]
\centering
\includegraphics[width=0.5\textwidth]{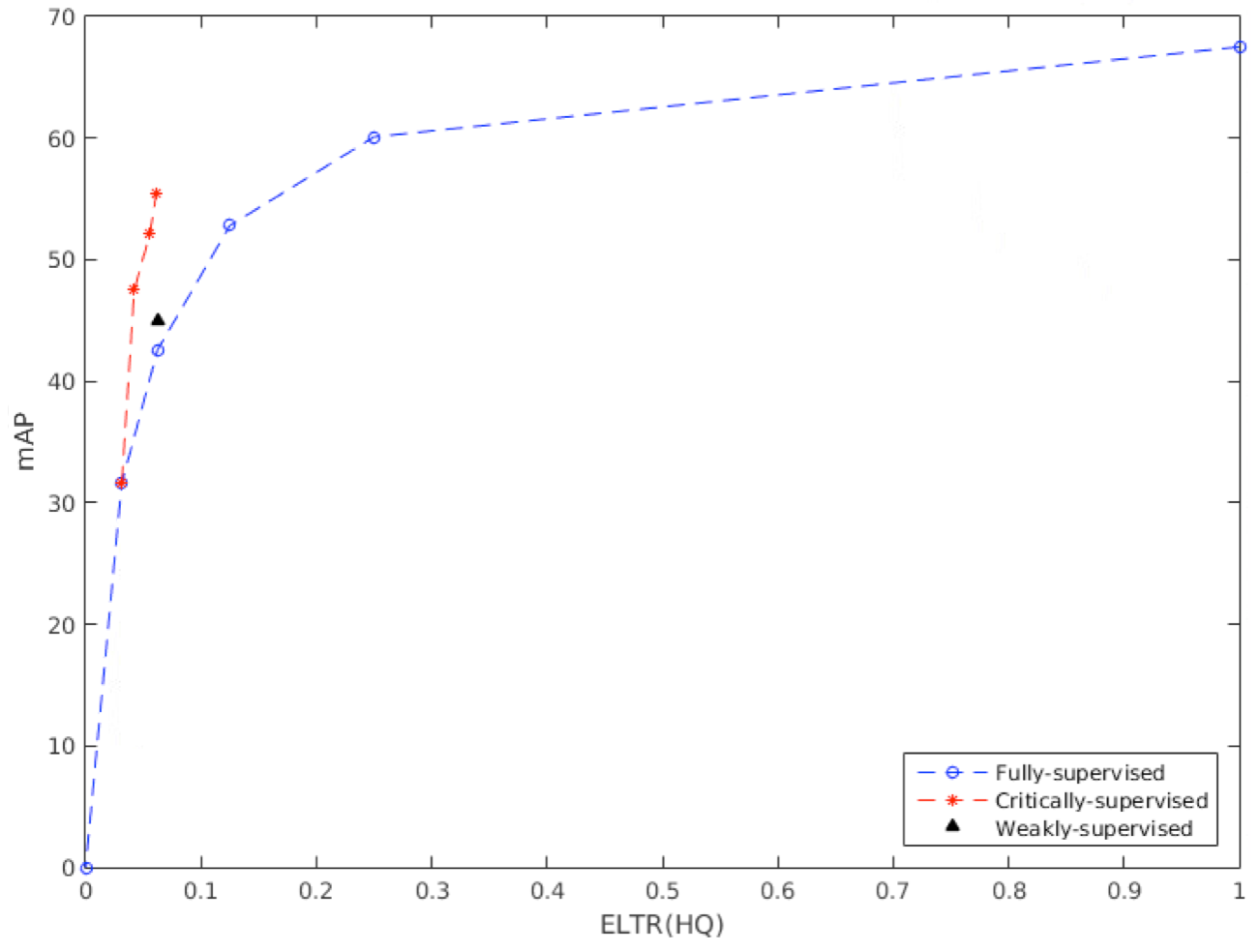}
\caption{The MAP-vs-ELTR performance comparison of fully,
weakly, and critically supervised learning on the VOC07 dataset,
where each dot denotes a model trained using different training sets 
with the estimated labeling time.}\label{fig.TOA_voc07_curve}
\end{figure}

We validate the TOA method on the VOC07 dataset in this subsection.  To
compare critical, full and weak supervision methods, we show a
mAP-vs-ELTR plot in Fig. \ref{fig.TOA_voc07_curve}, where the HQ profile
is used for ELTR calculation. In the figure, the points on the blue
curve are trained using full supervision with different sizes of image
subsets. The black point of triangle shape on the curve is the
state-of-the-art weak supervision performance. Under the same mAP, its
labeling time is less than that of full supervision. The points on the
red curve show the performance of the TOA method in different stages.
The TOA method has a higher mAP than the weak supervision method.  It
also demands much less labeling time than full and weak supervision
methods. 

In this experiment, our $S_0$ model uses only $157$ images, which
are $1/64$ of the VOC07 trainval set. Then, different
QA numbers are conducted in each stage, as detailed in Table
\ref{tab.TOA_voc07overall}. For example, in stage-$1$, 3600 QAs are done
and the mAP of $S_1$ boosts from 32.0\% to 47.5\% with only 0.0421 ELTR
required. In stage-$2$, an additional 3600 QAs are performed. A total
of 7200 QAs contribute to a mAP of 52.2\% with 0.0559 ELTR. In
stage-$2$, some QAs are of type-2 which are decided by the CEM
algorithm. 

\begin{table}[h!]\centering
\caption{The experimental setup for Fig. \ref{fig.TOA_voc07_curve},
where the numbers in parenthesis after FRCNN means the percentage of
data used for labeling. The parenthesis after TOA means the number of
stages used in TOA. In this table, the numbers of full-labeled images
and QA-labeled samples are provided along with the final estimated
labeling time ratio (ELTR).} \label{tab.TOA_voc07overall}
\resizebox{0.5\textwidth}{!}{
\begin{tabular}{|l|l|l|l|l|l|l|l|}
\hline
method   & Category & mAP & $N^{Full}$ & $N^{QA1}$ & $N^{QA2}$ & $N^{QA}$ & ELTR(HQ) \\ \hline
FRCNN    & Full supervision   & 67.5 & 5011 &   0  &   0  &  0  &  1 \\ \hline
FRCNN(1/4)  & Full supervision & 60.1 & 1253 &   0  &   0  &  0  &  0.25 \\ \hline
FRCNN(1/8)  & Full supervision & 52.8 &  626 &   0  &   0  &  0  &  0.125 \\ \hline
FRCNN(1/16) & Full supervision & 42.6 &  313 &   0  &   0  &  0  &  0.0625 \\ \hline
FRCNN(1/32) & Full supervision & 32.0 &  157 &   0  &   0  &  0  &  0.0313 \\ \hline
K-EM\cite{yan2017weakly} & Weak supervision  & 46.1 &   0 &   0  &   0  &  0  &  0.0619 \\ \hline
TOA(S1)    & Critical supervision & 47.5 &  157 &   3600  &   0  &  3600  &  0.0421 \\ \hline
TOA(S2)    & Critical supervision & 52.2  &  157 &   5276  &  1924  & 7200  &  0.0559 \\ \hline
TOA(S3)    & Critical supervision & 55.5 &  157 &   5276  &  3124  &  8400  &  0.0613 \\ \hline
\end{tabular}}
\end{table}

The benefits of the TOA method can be examined from two aspects.  For
the same labeling time with ELTR equal to 0.062, the TOA can achieve a
55.5\% mAP, which out-performs K-EM (weakly supervised method) by 9.4\%,
and exceeds FRCNN by 12.9\%. On the other hand, under the same mAP of
55\%, the labeling time needed for the TOA method is only about one
third of the labeling time of the FRCNN while weak supervision cannot
reach this level. The detection performance on different classes is
shown in Table \ref{tab.TOA_voc07overall_byclass}. The TOA method can
achieve consistent improvement across all object classes. 

\begin{table*}[h!]\centering
\caption{Comparison of detection results for each object class with the
experimental setup given in Table \ref{tab.TOA_voc07overall}, and
visualized in Fig. \ref{fig.TOA_voc07_curve}} 
\label{tab.TOA_voc07overall_byclass}
\resizebox{1\textwidth}{!}{
\begin{tabular}{l|llllllllllllllllllll|l}
method  & aero & bike & bird & boat & bottle & bus & car & cat & chair & cow & table & dog & horse & mbike & person & plant & sheep & sofa & train & tv & mAP \\ \hline
FRCNN(1/32)  & 24.8 & 27.4 & 30.2 & 11.7 & 14.8 & 34.4 & 56.9 & 44.4 & 18.3 & 43.8 & 17.7 & 37.7 & 52.4 & 45.3 & 51.2 & 11.2 & 32.0 & 24.9 & 23.4 & 36.7 & 32.0     \\ 
FRCNN(1/16)  & 39.9 & 58.2 & 42.6 & 28.0 & 15.8 & 50.3 & 66.4 & 59.1 & 29.2 & 46.4 & 30.4 & 44.8 & 54.3 & 62.6 & 54.5 & 14.3 & 24.0 & 32.9 & 43.9 & 53.9 & 42.6 \\ 
FRCNN(1/8)   & 47.4 & 65.4 & 50.0 & 34.4 & 24.9 & 65.6 & 71.7 & 69.0 & 28.3 & 62.9 & 41.2 & 64.3 & 70.2 & 62.6 & 61.8 & 23.3 & 59.6 & 51.8 & 53.1 & 49.1 &  52.8 \\ 
FRCNN(1/4)   & 58.8 & 70.4 & 58.1 & 42.9 & 43.6 & 70.7 & 75.9 & 76.2 & 34.4 & 66.3 & 52.4 & 69.5 & 76.4 & 73.6 & 67.5 & 32.6 & 59.3 & 56.6 & 63.7 & 52.6 &  60.1 \\ 
FRCNN    & 68.6 & 77.8 & 65.0 & 57.5 & 44.8 & 79.9 & 78.6 & 83.8 & 41.6 & 74.5 & 62.0 & 82.3 & 80.8 & 75.3 & 70.8 & 37.7 & 66.2 & 68.0 & 75.4 & 60.0 &   \textbf{67.5}    \\ \hline
K-EM  & 59.8 & 64.6 & 47.8 & 28.8 & 21.4 & 67.7 & 70.3 & 61.2 & 17.2 & 51.5 & 34.0 & 42.3 & 48.8 & 65.9 & 9.3 & 21.1 & 53.6 & 51.4 & 54.7 & 50.7 &  \textbf{46.1} \\ \hline
TOA(S1)  & 48.1 & 61.5 & 43.6 & 22.5 & 29.6 & 60.8 & 61.7 & 56.1 & 24.8 & 57.3 & 28.4 & 56.4 & 63.7 & 62.9 & 47.7 & 21.5 & 47.8 & 45.6 & 55.9 & 54.5 &  47.5 \\ 
TOA(S2)  & 53.6 & 48.3 & 55.3 & 31.0 & 27.2 & 61.1 & 66.9 & 70.2 & 26.1 & 68.8 & 32.3 & 66.8 & 71.1 & 62.8 & 57.2 & 21.8 & 54.0 & 52.0 & 64.5 & 52.1 &  52.2 \\ 
TOA(S3) & 60.0 & 52.8 & 57.6 & 35.3 & 29.6 & 66.3 & 69.2 & 73.2 & 29.5 & 70.3 & 42.5 & 66.5 & 72.8 & 66.2 & 58.4 & 26.2 & 57.5 & 54.3 & 65.4 & 57.2 &  \textbf{55.5} \\ 
\end{tabular}
}
\end{table*}

Besides the VOC07 dataset, we apply the TOA method to the VOC07+12
trainval dataset, and test it on the VOC07 test set. Since the number of
images becomes larger, the setup of the TOA method is different and it
is given in Table \ref{tab.TOA_voc12overall}. Similar performance gains
can be observed, and a higher mAP of 58.3\% is achieved with an even
lower ELTR. 

\begin{table}[h!]\centering
\caption{The experimental setup for the fully and critically supervised
CNNs trained on the VOC07+12 trainval dataset and tested on the VOC07
dataset. The numbers of full-labeled images and QA-labeled samples are
provided along with the final estimated labeling time ratio (ELTR).}
\label{tab.TOA_voc12overall}
\resizebox{0.5\textwidth}{!}{
\begin{tabular}{|l|l|l|l|l|l|l|l|}
\hline
method   & Category & mAP & $N^{Full}$ & $N^{QA1}$ & $N^{QA2}$ & $N^{QA}$ & ELTR(HQ) \\ \hline
FRCNN    & Full supervision   & 70.7 & 16551 &   0  &   0  &  0  &  1 \\ \hline
FRCNN(1/4)  & Full supervision & 66.8 & 4138 &   0  &   0  &  0  &  0.25 \\ \hline
FRCNN(1/8)  & Full supervision & 61.6 & 2069 &   0  &   0  &  0  &  0.125 \\ \hline
FRCNN(1/16) & Full supervision & 56.1 & 1034 &   0  &   0  &  0  &  0.0625 \\ \hline
FRCNN(1/32) & Full supervision & 49.0 &  517 &   0  &   0  &  0  &  0.0313 \\ \hline
FRCNN(1/64) & Full supervision & 38.3 &  259 &   0  &   0  &  0  &  0.0156 \\ \hline
TOA(S1)    & Critical supervision & 54.6 &  259 &   10000  &   0  &  10000  &  0.0248 \\ \hline
TOA(S2)    & Critical supervision & 58.3  &  259 &   16038  &  3962  & 20000  &  0.0357 \\ \hline
\end{tabular}}
\end{table}

\subsection{Single-Class Object Detection}\label{sec.TOA_Caltech}

\begin{figure}[h!]
\centering
\includegraphics[width=0.50\textwidth]{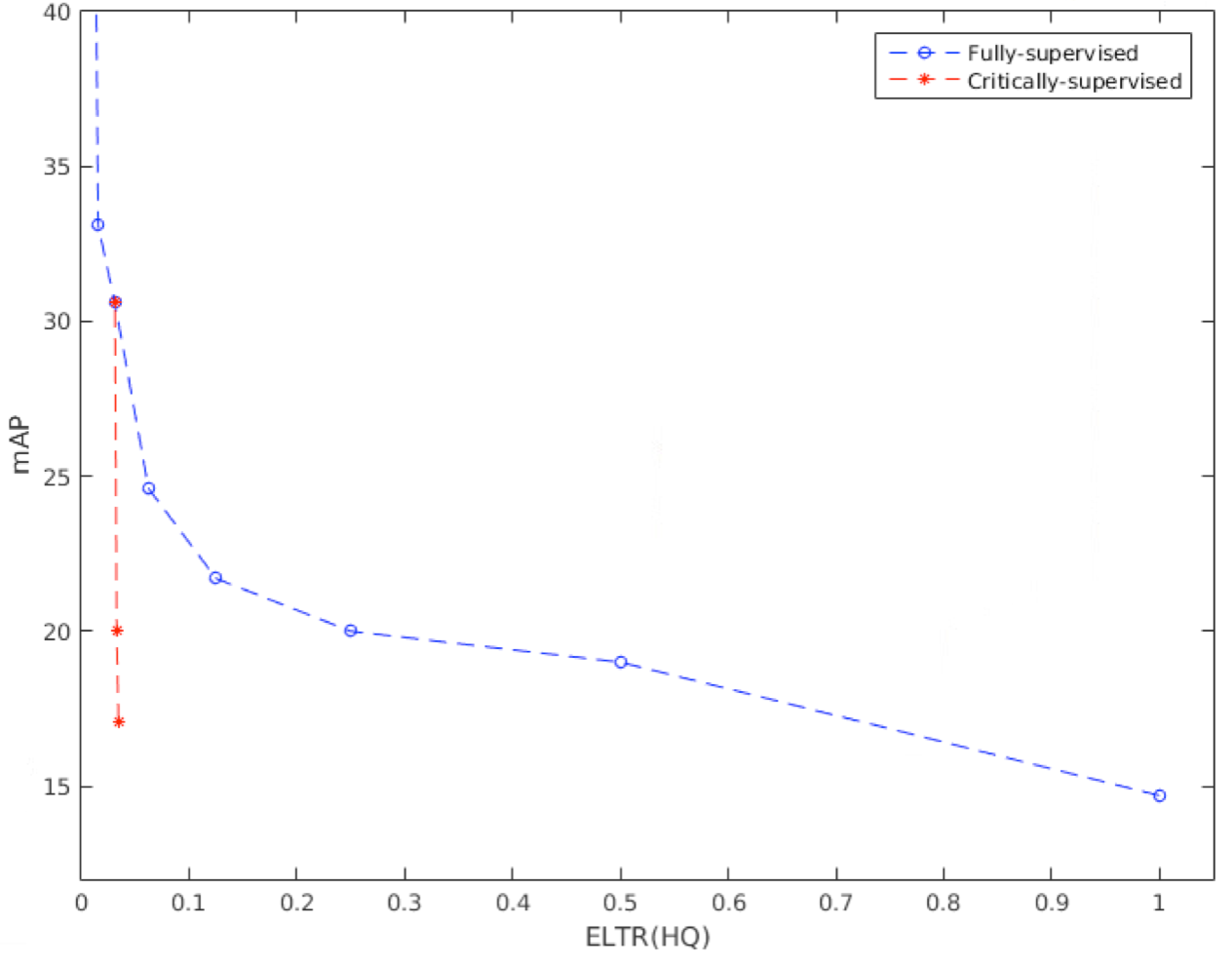}
\caption{The MR-vs-ELTR performance comparison of fully and critically
supervised learning on the Caltech dataset, where each dot denotes a
model trained under different setups with their associated labeling
time.}\label{fig.TOA_Caltech_curve}
\end{figure}

In this subsection, we evaluate the TOA method on the single object
detection problem using the Caltech dataset. The standard evaluation
metric for the Caltech dataset is the miss rate (MR) versus the
false-positive-per-image (FFPI) curve, which is calculated for each
trained model. Then, the overall performance of a model is obtained by
averaging several reference points on the curve in log scale. We use the
MR to represent the log-average-miss-rate, and compare the performance
of the TOA method with the full supervision scheme in Fig.
\ref{fig.TOA_Caltech_curve}. In these experiments, we implement the
FRCNN adapted to the pedestrian detection problem according to
\cite{li2015scale}.  We see from the figure that the performance gain is
even more obvious for the Caltech dataset. For example, when $MR=17$,
the TOA can save up to 95\% of the labeling efforts as compared to that
of the full supervision method. This is attributed to fact that the
Caltech dataset contains more redundant samples than the VOC datasets.
The Caltech dataset is captured frame-by-frame with street-view scenes,
and most pedestrians have similar shapes and appearances.  Clearly, the
TOA method has an advantage when a dataset has more similar (or
redundant) samples. 

\begin{figure}[h!]
\centering
\includegraphics[width=0.50\textwidth]{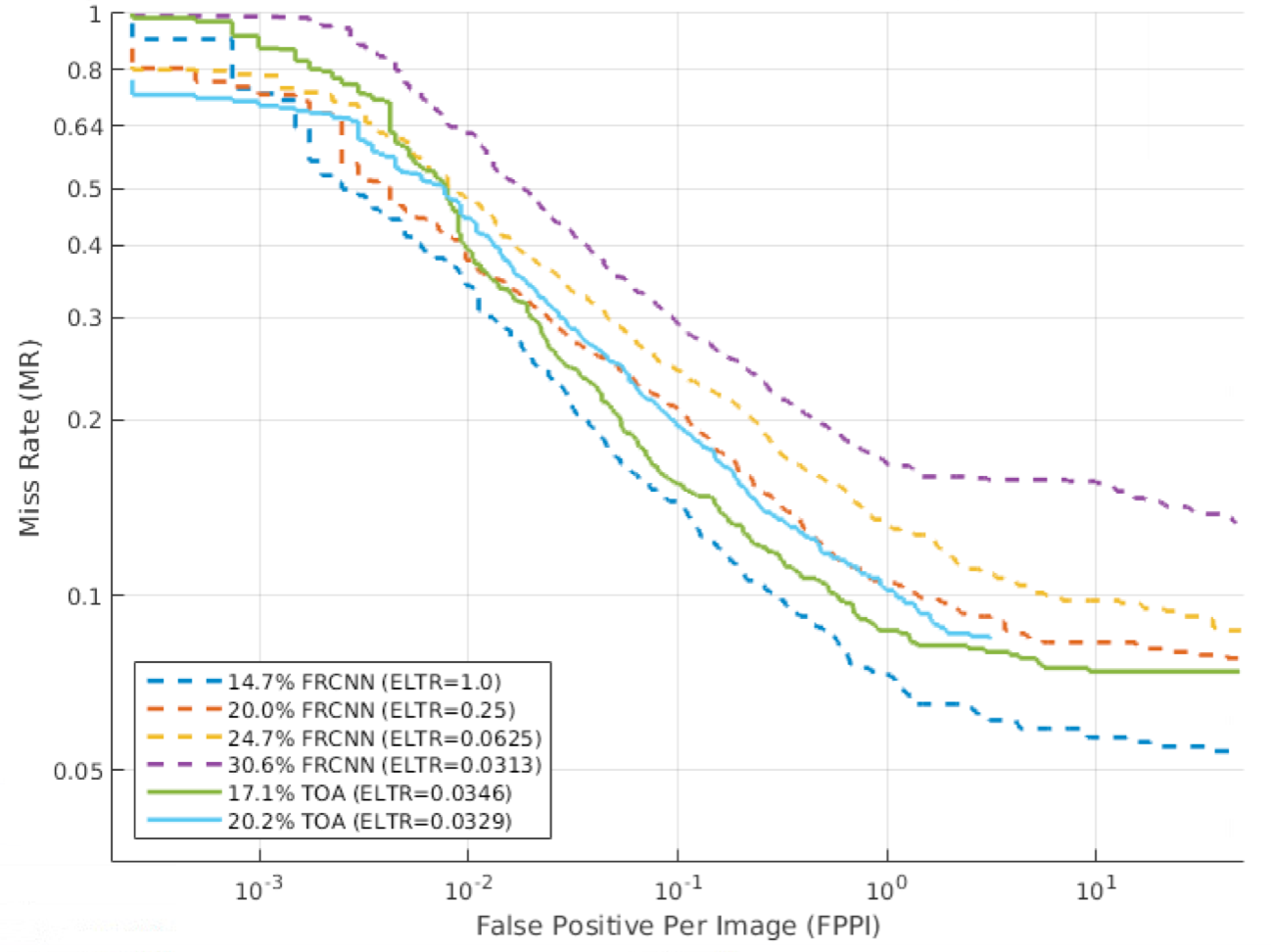}
\caption{The MR-vs-FPPI performance curves for models defined in Table
\ref{tab.TOA_Caltech_overall}.}\label{fig.TOA_MR_FPPI}
\end{figure}

The detailed setup of our experiment is given in Table
\ref{tab.TOA_Caltech_overall}, where each entry gives a dot on the the
MR-vs-ELTR curve. The MR-vs-FPPI (false-positive-per-image) curves of
these models are shown in Fig. \ref{fig.TOA_MR_FPPI}.  Although the
Caltech dataset provides frame-to-frame images, we use one image every 4
frames to form our baseline dataset $D$. This is because that the
computational time can be saved dramatically during the whole TOA
process, and the performance with one quarter of training data is not
much different from that of using every frame as shown in Table
\ref{tab.TOA_Caltech_overall}. 

\begin{table}[h!]\centering
\caption{The experimental setup for Fig.  \ref{fig.TOA_Caltech_curve},
where the TOA method is applied to a subset(1/4) of the Caltech
dataset so that the ELTR for FRCNN(1/4) is equal to one.}
\label{tab.TOA_Caltech_overall}
\resizebox{0.5\textwidth}{!}{
\begin{tabular}{|l|l|l|l|l|l|l|} \hline
           & MR & $N^{Full}$ & $N^{QA1}$ & $N^{QA2}$ & $N^{QA}$ & ELTR(HQ) \\ \hline
FRCNN      & 14.4 &  128419  &  0   &  0   &      0      &      \\ \hline
FRCNN(1/4) & 14.7 &  32105 &   0  &   0  &      0      &   1   \\ \hline
FRCNN(1/8) & 19.0 &  16052 &  0  &   0  &       0     &    0.5  \\ \hline
FRCNN(1/16) & 20.0  & 8026 &  0   &  0   &      0      &   0.25  \\ \hline
FRCNN(1/32) & 21.7  & 4013 &  0   &  0   &      0      &   0.125   \\ \hline
FRCNN(1/64) & 24.6  & 2007 &  0   &   0  &       0     &   0.0625   \\ \hline
FRCNN(1/128) & 30.6  & 1003 &   0  &  0   &      0      &  0.0313    \\ \hline
TOA(S1) & 20.2 &  1003 &    50000 &  0   &   5000    &   0.0329 \\ \hline
TOA(S2) & 17.1 &  1003 &  10000   &  0   &  10000    &  0.0346    \\ \hline
\end{tabular}}
\end{table}

\begin{figure}[h!]
\centering
\includegraphics[width=0.5\textwidth]{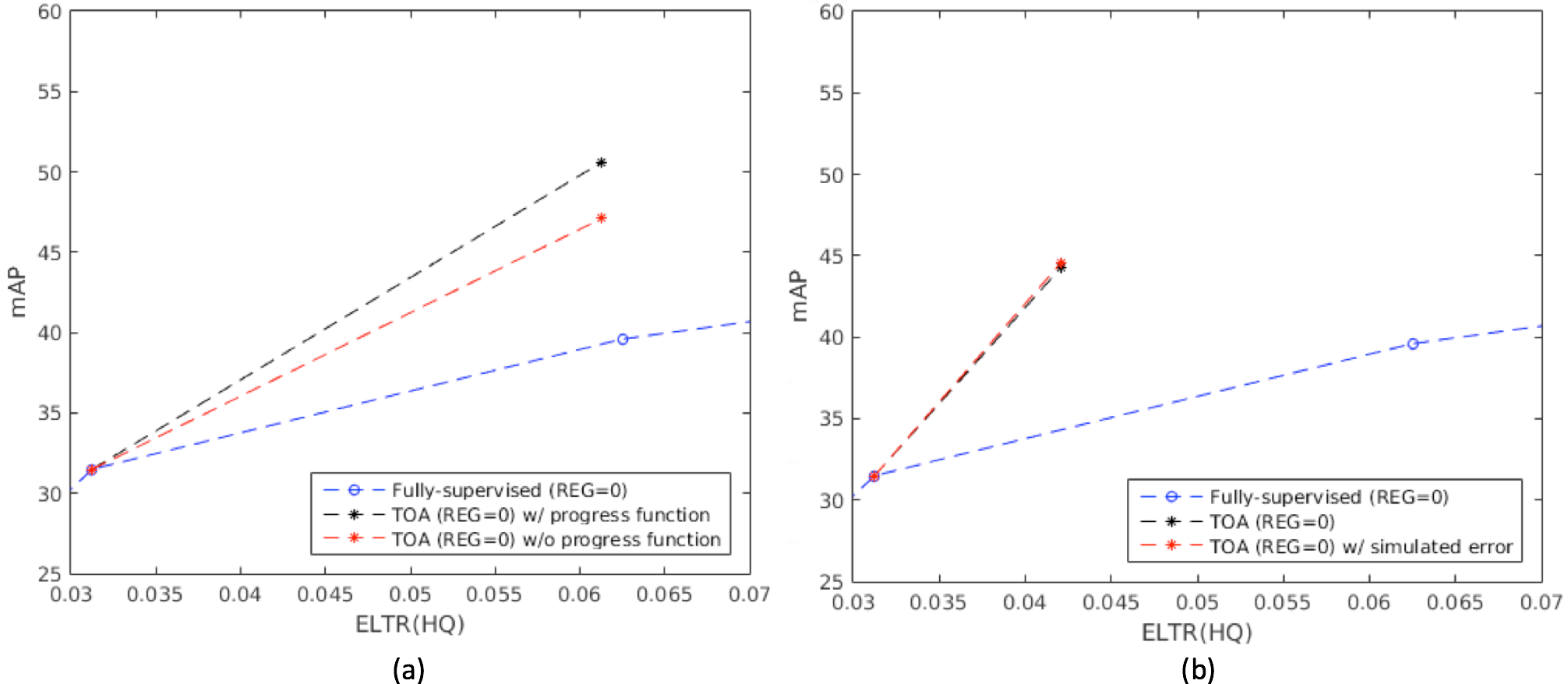}
\caption{Analysis results for the TOA method: (a) the effectiveness of the
progress function for hard samples, and (b) the results by taking human 
errors into account in the simulation.}\label{fig.TOA_analysis}
\end{figure}

\subsection{Analysis on CEM, NOP and Labeling Time Models}\label{sec.TOA_analysis}

We conduct further analysis on CEM, NOP and labeling time models 
on the VOC07 dataset to gain more insights into the TOA method.

\noindent
{\bf Effectiveness of CEM.} We attempt to explain the proposed CEM
algorithm. First, we justify the use of detection scores in the
criticalness function. The valid QA rates are provided in Table
\ref{tab.TOA_validQA}, which indicate the QA percentages that result
in labeled objects but background. As mentioned before, only object
samples can help improve the precision since we already have NOP
samples.  For the type-1 QA to be a valid one, the sample has to be an
object and the class should be correct. For the type-2 QA to be a valid
one, it requires the sample to be an object only.  Since the initial
model $S_0$ is a weak detector, the valid QA-1 rate is only around 37\%
in stage-$1$. Thus, it is important to use samples of high detection
score for the QA in early stages. Otherwise, more QAs will be wasted. 

\begin{table}[h!]\centering
\caption{The valid QA rate with the CEM algorithm in each stage.}
\label{tab.TOA_validQA}
\begin{tabular}{|l|l|l|l|l|}\hline
        & QA1 & Valid QA1 & QA2 & Valid QA2 \\ \hline
TOA(S1)    & 3600  & 1320 & 0 & 0 \\ \hline
TOA(S2)   &  5276 & 1754 & 1924 & 904 \\ \hline
TOA(S3)    & 5276 & 1754  & 3124 & 1406 \\ \hline
\end{tabular}
\end{table}

Second, we justify the class balancing component in the progress function.
We compare the original detection result TOA(S1) with a model that does
not have this progress function.  The class-wise detection results for
both models are shown in Table \ref{tab.TOA_balance}. Under the same QA
amount, there is a huge mAP difference of 5.3\%. The progress function
can help select balanced training samples among different classes to
achieve performance boost in multiple object categories. In contrast,
the model without the progress function tends to select redundant
samples from major classes such as the ``person" or the ``car".  It does
not find the most critical examples to boost the overall detection
performance, since samples of other object categories are not properly
represented. 

\begin{table*}[h!]\centering
\caption{Differences in detection results using models with/without the
class-balancing progress function.} \label{tab.TOA_balance}
\resizebox{1\textwidth}{!}{
\begin{tabular}{l|l|llllllllllllllllllll|l|}
method & train set & aero & bike & bird & boat & bottle & bus & car & cat & chair & cow & table & dog & horse & mbike & person & plant & sheep & sofa & train & tv & mAP \\ \hline
TOA(S1)  &  07 & 42.9 & \textbf{61.5} & 40.9 & \textbf{19.0} & \textbf{24.6} & 55.3 & 59.2 & 55.0 & 21.7 & 56.0 & \textbf{33.8} & 47.4 & 60.1 & 58.6 & 43.0 & 21.9 & \textbf{45.4} & 41.9 & \textbf{52.6} & \textbf{44.7} & 44.3 \\ 
TOA(S1) w/o balance &  07 & 41.5 & 25.2 &  41.7 & 6.2 & 18.8 & 54.0 & \textbf{62.7} & 52.0 & 19.5 & 49.9 & 20.9 & 48.9 & 58.4 & 54.2 & \textbf{54.8} & 15.9 & 30.7 & 41.3 & 44.4 & 38.9 & 39.0 \\ 
\end{tabular}}
\end{table*}

Third, we explain the hardness component in the progress function. By
removing this component from the criticalness function, we show the
result in Fig. \ref{fig.TOA_analysis}(a), which has worse mAP
performance under the same experimental setup using TOA(S3).  It
indicates that the hardness component plays an important role in
highlighting different types of critical examples in different labeling
stages. 

Finally, we discuss the value of the representativeness component in the
progress function. It is worthwhile to compare redundancy of samples in
the VOC and the Caltech datasets. By examining the kNN graphs for both
datasets, we find that the averaged Euclidian distance between two
nearest samples is 249.3 and 1194.2 for the Caltech and the VO07
datasets, respectively. In other words, samples in the Caltech dataset
are closer than those in the VOC07 dataset. Based on this observation,
we set $\delta=0.2$ for the VOC dataset and $\delta=1$ for the Caltech
dataset, respectively, in adaptation to their different properties. 

\noindent
{\bf Effectiveness of NOP.} We examine the quality of the NOP and the
whole training samples. The quality of the NOP can be checked by
calculating its precision of retrieving true negative samples. In the
VOC07 dataset, the NOP algorithm can select an average of 1871 bounding
boxes out of 1900 object proposals per images, meaning that a large
percentage of boxes are still kept under the InvSet operation. For the
whole NOP set, the negative sample precision is as high as 99.99914\%.
This is better than the precision of around 99.5\% using alternative
approaches as shown in Fig.  \ref{fig.TOA_nop_failure}. Recall that the
NOP precision should be extremely high to avoid the network being confused by false negative. 

\begin{figure}[h!]
\centering
\includegraphics[width=0.5\textwidth]{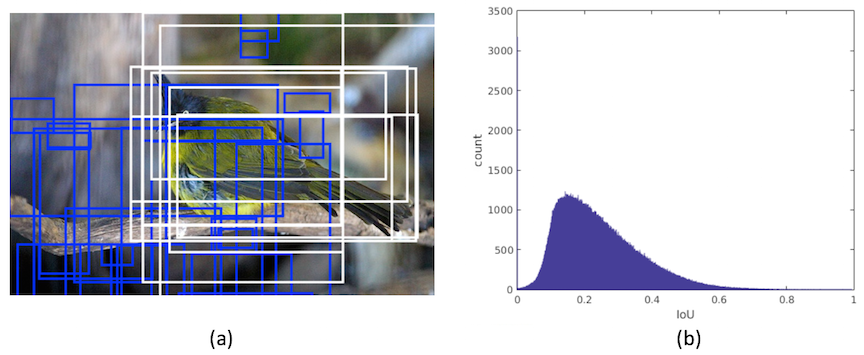}
\caption{Effectiveness of the NOP: (a) examples of training sample composition with 50 boxes,
and (b) the IoU distribution of all negative samples as compared with the
ground truth.}\label{fig.TOA_composition}
\end{figure}

The training samples in TOA include both QA-labeled samples and NOP samples. An
example is shown in Fig. \ref{fig.TOA_composition}(a). In general, the
visual quality of these training samples are high, where the NOP samples
contain informative negative samples with different sizes.  We also show
the IoU histogram, which compares all NOP samples with the ground truth
on the VOC07 dataset, in Fig. \ref{fig.TOA_composition}(b).  Ideally, the
IoU should fall in the range of $TH_{HI},TH_{LO}$ with full supervision.
For the TOA method, the IOU values of the majority of negative samples
are in the range. 

\noindent
{\bf Simulation Variation.} We simulate human QA behavior in
experiments. Humans can make inaccurate decision.  Even if there is no
intentional error, humans cannot determine objects based on a certain
IoU threshold easily. For this reason, we add some random errors in the
simulation. The original design is to ask humans to select objects with
an IoU greater than 0.6. Here, we use a uniform random variable in the
range of [0.6 + $\psi$, 0.6 - $\psi$] to simulate human thesholding. We
set $\psi=0.1$ and plot the results in Fig.  \ref{fig.TOA_analysis}(b).
We see from the figure that the TOA method is robust against small
labeling variations. 

\begin{figure}[h!]
\centering
\includegraphics[width=0.5\textwidth]{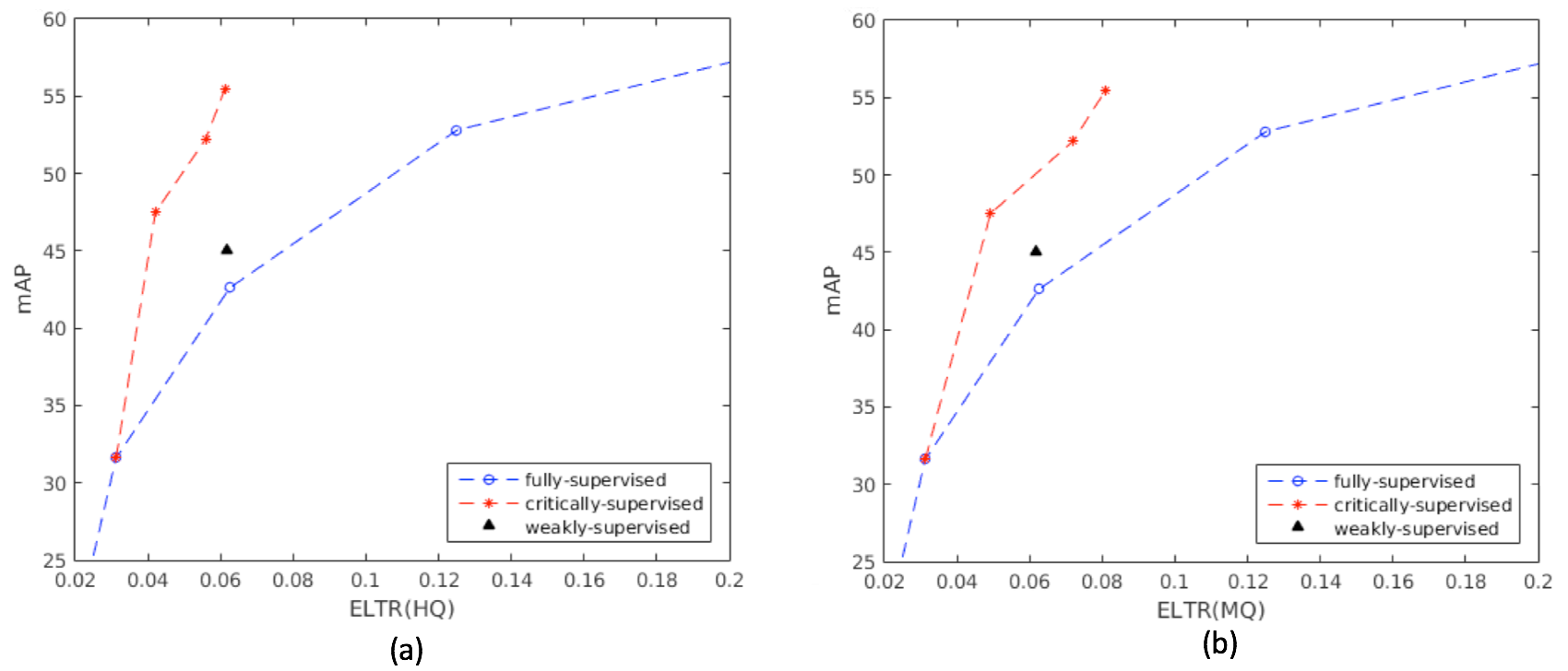}
\caption{Experimental results with different labeling time models: (a)
the HQ profile and (b) the MQ profile.}\label{fig.TOA_label_time}
\end{figure}

\noindent
{\bf Labeling Time Model.} Labeling time can vary according to the
labeling tools.  We apply the HQ profile time model that adopts numbers
from several papers to provide an accurate estimation in the crowd
sourcing setup. We also evaluate the time model using the MQ profile,
which is in favor of the fully supervised learning. By following the
experimental setup in Sec. \ref{sec.ch5_labeltime}, we show the results
for the VOC07 dataset in Fig.  \ref{fig.TOA_label_time}(b).  We still see clear
performance gains of the TOA method over the full supervised one. 

\subsection{Other Training Details}\label{sec.TOA_details}

Being different from the traditional full supervision training, the TOA
method has different types of training samples and follows an adaptive
procedure in training the model. In this subsection, we briefly discuss how
various training options affect the final results, and explain about our designed training details.
Since the performance in Caltech dataset is superior and thus not sensitive to
different training setting, we focus on the more complex VOC07 dataset.

\begin{figure*}[h!]
\centering
\includegraphics[width=1\textwidth]{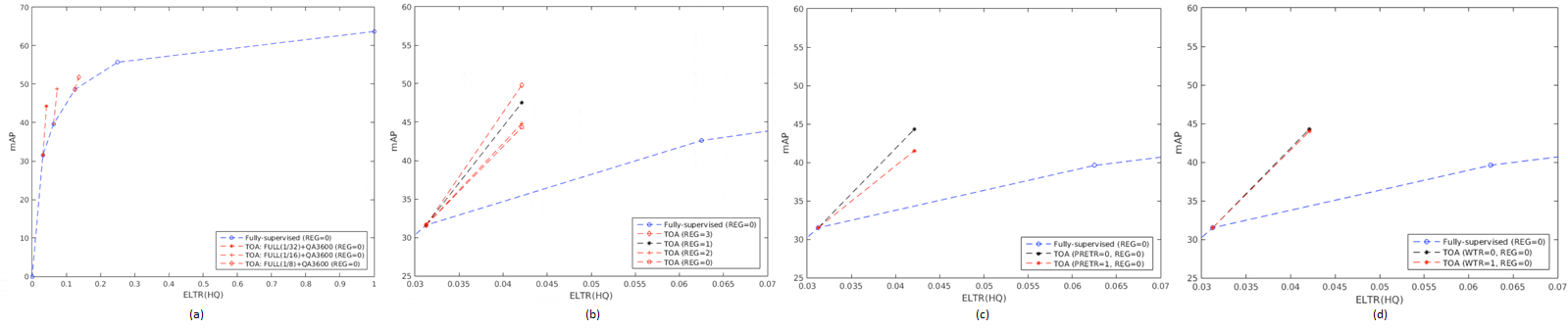}
\caption{Experimental results with different training parameters: (a)
the size of the initial set, (b) bounding box regression (REG), (c) pretrained model selection
(PRETR), (d) weighing on
image subsets (WTR).  Results are provided for a zoomed-in ELTR range except
for (a).} \label{fig.TOA_train_options}
\end{figure*}

\noindent
{\bf Labeling Amount Selection.} The selection of the labeling amount,
including the number of initial fully labeled image subset and the
number of QA in each stage, is heuristic.  It depends on the
availability of annotation resources. Generally speaking, if a larger
initial set is selected, detection scores and deep features are more
reliable and better QA samples can be retrieved more efficiently.
However, full-labeling is time consuming, leading to fast growth in
labeling time. In contrast, if the initial set is too small, the poor
$S_0$ detector will have bad performance in selecting CEM samples.  In
Fig. \ref{fig.TOA_train_options}(a), we show the results with different
initial sets, and then apply the same number of QA. In general, a
smaller initial set is still favorable if the $S_0$ detector is not too
poor. This is because QA-labeling consumes much less labeling time than
full-labeling, and this choice will be more efficient in the early
stage.

\noindent
{\bf Bounding Box Regression (REG).} Object detection involves
multi-task training of probability scoring and bounding box regression
simultaneously \cite{girshick2015fast}. For bounding box regression, the
ground truth bounding box locations are used as training targets.
However, in the TOA method, we do not have ground truth bounding box
locations. This limitation can be handled in several ways. The first one
is to train the model without bounding box regression at all (with
REG=0), which gives the worst performance in Fig.
\ref{fig.TOA_train_options}(b). An alternative solution is to accept the
locations of these noisy bounding boxes as the groud truth with REG=1.
We see from the figure that training the regressor with noisy bounding
boxes still provide reasonable performance improvement. The result is
better than that without any regression or using the weak regressor
trained from the $S_0$ detector (REG=2). We also show the performance of
the ideal case with an optimal regressor (REG=3), which is trained using
the VOC2007 trainval set, in the figure for performance benchmarking.
Our work emphasizes better on the classification performance. If accurate localization \cite{papadopoulos2016we} is
desired, a better regressor can be further added to the TOA method. 
In Sec. \ref{sec.TOA_VOC} and Sec. \ref{sec.TOA_Caltech}, we adopt REG=1, and for rest of 
the experiments we do not adopt bounding box regression for analysis purposes.  

\noindent
{\bf Pretrained Model Selection (PRETR).} For each stage $n$, training
samples are collected from both fully labeled and QA-labeled subsets.
The training data set in stage $n$ is always a superset of that in the
previous stage. To retrain CNN model $S_n$, we have two choices: 1)
using the ImageNet pretrained model (PRETR=0) as done in stage $S_0$, or
2) applying the pretrained model from the previous stage and do
fine-tuning (PRETR=1). For the latter, the learning rate can be reduced
since it is built upon a better reference.  The result is shown in Fig.
\ref{fig.TOA_train_options}(c). It shows that the directly re-training on 
ImageNet pretrained features results in better performances,
which helps avoid bad feedback loop from previous training stages.

\noindent
{\bf Weighing on Image Subsets (WTR).} The TOA training set consists of
both fully-labeled and QA-labeled subsets.  The fully-labeled subset
contain training samples of higher quality than the QA-labeled subset.
It could be advantageous to give the fully-labeled frames higher weights
in the training process. This can be achieved by allowing the fully
supervised frames to appear more times in each epoch (WTR=1).  We show
the result of weighing fully labeled images two times in Fig.
\ref{fig.TOA_train_options}(d).  The weighting scheme does not provide
better performance. It implies that the quality of training samples
prepared by the TOA method is already good. 

\section{Conclusion}\label{sec.TOA_conclude}

A critically supervised learning methodology for object detection was
studied in this work with an objective to maintain good detection
performance yet with significantly less labeling effort as compared with
full supervision.  Specifically, we proposed the TOA method that consists
of several novel components. It used the CEM algorithm to select sampels
for QA and the NOP to extract negative samples without extra labeling.
The effectiveness of the TOA method was demonstrated on the VOC dataset
and the Caltech pedestrian dataset. Extensive experiments were conducted
to provide insights into the TOA method.

\bibliographystyle{IEEEtran}
\bibliography{egbib}

\end{document}